%% file: RetryRoute.tex
\documentclass[sigconf,nonacm]{acmart}
\setcopyright{none}
\acmConference[WWW '27]{The ACM Web Conference 2027}{May 10--14, 2027}{Dublin, Ireland}
\usepackage{booktabs}
\usepackage{multirow}
\usepackage{xspace}
\usepackage{amsmath}
\usepackage{graphicx}
\usepackage{subcaption}
\newcommand{\sys}{RetryRoute\xspace}
\newcommand{\sticky}{sticky failures\xspace}
\newcommand{\Sticky}{Sticky failures\xspace}
\newtheorem{proposition}{Proposition}
\newtheorem{lemma}{Lemma}

\begin{document}

\title{RetryRoute: Retry-Aware Model Routing for Agentic Image Editing Services}

\author{Hanshuai Cui}
\affiliation{\institution{School of Artificial Intelligence, Beijing Normal University, Beijing 100875, China}\city{}\country{}}
\affiliation{\institution{Institute of Artificial Intelligence and Future Networks, Beijing Normal University, Zhuhai 519087, China}\city{}\country{}}
\author{Zhiqing Tang}
\affiliation{\institution{Institute of Artificial Intelligence and Future Networks, Beijing Normal University, Zhuhai 519087, China}\city{}\country{}}
\author{Zhi Yao}
\affiliation{\institution{School of Artificial Intelligence, Beijing Normal University, Beijing 100875, China}\city{}\country{}}
\affiliation{\institution{Institute of Artificial Intelligence and Future Networks, Beijing Normal University, Zhuhai 519087, China}\city{}\country{}}
\author{Jun Luo}
\affiliation{\institution{College of Computing and Data Science, Nanyang Technological University, Singapore}\city{}\country{}}
\author{Weijia Jia}
\affiliation{\institution{Institute of Artificial Intelligence and Future Networks, Beijing Normal University, Zhuhai 519087, China}\city{}\country{}}

\begin{abstract}
Web agents increasingly edit images by calling open editing models in a critique-and-retry loop, and these models differ in cost by more than an order of magnitude.
After every rejected edit, a service must decide whether to retry the same model, escalate to another or give up.
Existing routers pick one model per request, and cascades escalate in a fixed order.
However, both treat a retry as an independent draw.
We identify \emph{\sticky}, meaning that a retry after a failure succeeds only 17--53\% as often as a first try across six open models, three benchmarks and two judges.
Based on this finding, we propose \sys, a retry-aware router that estimates each model's chance of success given the request's failure history.
It plans retries, escalations and abandonment over failure states and adds live queue waits to the cost, so the same plan becomes load-aware dispatch.
At the main budget, \sys's success under the in-loop judge exceeds the strongest of eleven prior baselines by 2.8 points on average over eight groups.
On a GPU serving testbed it raises the share of edits delivered within the service-level objective by 5.0--6.3 points over the best prior baseline at every load.
\end{abstract}


\maketitle

\begin{figure}[t]
\centering
\begin{subfigure}[t]{\linewidth}\centering\includegraphics[width=\linewidth]{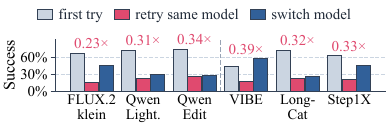}\caption{First try vs.\ retry after a failure, two judges averaged.}\end{subfigure}\\[3pt]
\begin{subfigure}[t]{\linewidth}\centering\includegraphics[width=\linewidth]{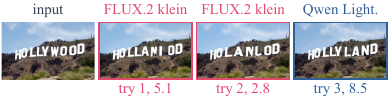}\caption{``Replace the text `WOOD' with `LAND'.''}\end{subfigure}
\caption{\Sticky. A model that fails an edit mostly fails it again, while a different model often succeeds.}
\Description{Top: grouped bars; for each of six editing models the success rate of a retry after a failure is 0.23 to 0.39 times the first-try rate per model, while switching to another model of the pool succeeds more often than the retry. Bottom: a sign edit where FLUX.2 klein fails twice and Qwen Lightning succeeds.}
\label{fig:sticky}
\end{figure}

\section{Introduction}
\label{sec:intro}

Image editing has become a routine tool call for Web agents~\cite{react,jarvisevo}, from recoloring product photos to localizing the text on a banner.
Following the critique-and-revise pattern of agent systems~\cite{selfrefine,reflexion,genartist,idea2img}, the agent sends the image and an instruction to an editing model, asks a vision-language judge whether the result follows the instruction, and retries when it does not.
The open editing models behind such calls~\cite{ldm,instructpix2pix} differ by more than an order of magnitude in cost.
A 4-step distilled model~\cite{progdistill} returns an edit in under 6\,s on a 24\,GB GPU, while the strongest open models need minutes.
However, a service that runs these loops pays for every rejected attempt, so where it sends the next attempt after a rejection decides both its cost and how many edits it delivers.
Retry routing is therefore a cost and reliability problem of agentic Web services, a topic of the Web Infrastructure and Agentic Systems track.

Serving systems for generative models already exploit this cost spread.
Per-request routers pick one model for each request from its features~\cite{rice,costaware,routellm}.
Following the cheap-first cascade of Viola and Jones~\cite{violajones}, serving cascades send a request to a light model first and escalate to a heavier one when a quality check fails~\cite{frugalgpt,automix,diffserve,hadis}.
Agent frameworks and Best-of-N sampling simply retry the same model with a new seed.
All three share an implicit assumption about the \emph{retry}.
They treat a second attempt on a failed model as an independent draw with the same success probability, and under this assumption retrying the cheapest adequate model is the cost-optimal policy (\S\ref{sec:why}).

However, these designs overlook how a failure changes the next attempt's chance of success, which concurrent work on code generation has begun to observe~\cite{resample}.
We measure it on six open editing models in two pools, on three benchmarks with GEdit-Bench in two languages, and with two independent judges, and identify \emph{\sticky}.
The retry rate is only 17--53\% of the first-try rate (Figure~\ref{fig:sticky}a).
FLUX.2 [klein]~\cite{flux2klein} passes 67\% of edits on the first try but only 15\% of the edits it has just failed, while another model of its pool still solves 45\% of them (Figure~\ref{fig:sticky}a).
This also explains why cascades help but stall.
They escalate after a failure, but in a fixed order and mostly once per model, so they rarely retry a cheap model while a retry is still worth its cost.

Turning \sticky into better retry decisions raises two challenges.
\textbf{(C1)} \emph{How can a service estimate each model's chance of success given a request's failure history?}
First-try rates overstate the retry rate by about two- to sixfold (\S\ref{sec:measure}), while counting outcomes per failure history spreads the logs thin, since few requests reach deep failure states.
\textbf{(C2)} \emph{How can it turn these estimates into retry, escalation and abandonment decisions that stay good as the load changes?}
When a request is hard for several models at once, a failure on one model also lowers the others' chances, so no fixed priority rule over the models is optimal in general.
A fixed retry plan tuned at one load can also collapse at another, because a busy GPU makes its model's next attempt slower (Figure~\ref{fig:motiv}c).

We therefore propose \sys, a router for agentic image editing services named for routing every retry rather than every request (Figure~\ref{fig:overview}).
To address C1, \sys estimates $p(a\mid s)$, the probability that model $a$'s next attempt succeeds given the failure counts $s$ so far, by counting logged attempts per failure state and shrinking each category's estimate toward the all-category one.
For C2, it solves a Bellman equation over failure states to plan retries, escalations and abandonment, adds each model's live queue wait to the cost and re-plans after every attempt, which turns escalation into load-aware dispatch.
We validate \sys against eleven prior baselines in eight evaluation groups and on a three-GPU serving testbed.
At the main budget and under the in-loop judge it leads the strongest prior baseline by 2.8 points on average and up to 5.6, and same-model retry by 3.5--13.3 (\S\ref{sec:offline}).
With one setting, its goodput on the testbed exceeds the best prior baseline's by 5.0--6.3 points at all three loads (\S\ref{sec:serving}).
Our contributions are threefold.
\begin{itemize}
\item We identify \sticky and show that retries of open editing models are strongly correlated with the failure they follow, across eight groups and 50 model--judge cells and even in the judges' late passes, and we give a formulation as planning over failure states that explains it.
\item We propose \sys, a router built on one estimate, the next attempt's success given the failure history, with an exact planner over failure states that charges live queue waits and an online learner on the same estimate.
\item We evaluate \sys against eleven prior baselines in eight offline groups and on a GPU serving testbed, with controls, ablations, robustness, generalization, per-category and judge checks.
\end{itemize}

\section{Related Work}
\label{sec:related}

\paragraph{Serving and routing generative image models.}
DiffServe~\cite{diffserve} and HADIS~\cite{hadis} cascade a light and a heavy text-to-image model under a quality check or a query-aware router, and Cost-Aware Routing~\cite{costaware} and GenRouter~\cite{genrouter} pick one model or one agentic workflow per request.
All of them decide once per request, whereas \sys plans across attempts.
OctoT2I~\cite{octot2i} chooses again after each rejected round. A large language model (LLM) picks the cheapest text-to-image model it expects to pass, from per-model success records and the earlier rounds. However, OctoT2I neither counts how a failure changes each model's next success nor plans the remaining attempts under a budget and load.

\paragraph{LLM routing and cascades.}
RouteLLM~\cite{routellm} routes each query between a strong and a weak model, FrugalGPT~\cite{frugalgpt} and AutoMix~\cite{automix} escalate when a scorer or a self-check rejects an answer, and TREACLE~\cite{treacle} learns when to re-query or switch LLMs under a budget.
Concurrent work on code generation finds that escalating beats resampling after a verifier rejection~\cite{resample}.
None of them, nor the algorithm selection view of routing~\cite{rice}, plans over failure states as \sys does.

\paragraph{Retries, sampling and agents.}
ACQB~\cite{acqb} learns from user retrials but gives a retrial the success probability of a first attempt, latency-aware routing~\cite{latencyroute} adds instance load to the first decision, and vLLM Semantic Router~\cite{vllmsr} switches models from recent outcomes without a planning model.
Agents refine their outputs from feedback~\cite{selfrefine} and editing agents revise them with tools~\cite{jarvisevo,genartist,idea2img}, and judges~\cite{llmjudge,viescore,editscore,editreward,edival} and hard-edit benchmarks~\cite{risebench} make such loops measurable. \sys instead decides which model the loop calls next and charges queue waits at every attempt.

\section{Background and Motivation}
\label{sec:prelim}

\begin{figure}[t]
\centering
\begin{subfigure}[t]{0.506\linewidth}\centering\includegraphics[width=\linewidth]{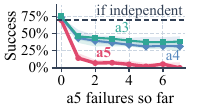}\caption{Next attempt after $f$ failures.}\end{subfigure}\hfill
\begin{subfigure}[t]{0.483\linewidth}\centering\includegraphics[width=\linewidth]{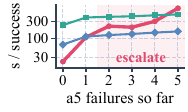}\caption{Cost per success.}\end{subfigure}\\[1pt]
\begin{subfigure}[t]{0.506\linewidth}\centering\includegraphics[width=\linewidth]{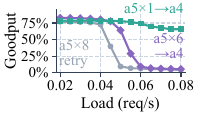}\caption{Fixed plans under load.}\end{subfigure}\hfill
\begin{subfigure}[t]{0.483\linewidth}\centering\includegraphics[width=\linewidth]{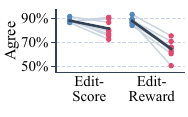}\caption{Single-judge check.}\end{subfigure}
\caption{Four findings behind \sys. Panels a and b use English GEdit-Bench with bands over four pool-1 groups, and panel c replays the logged edits under load on 3 GPUs.}
\Description{Four panels: the retry pass rate of the cheap model collapses after a failure while other models stay near 30 to 45 percent; its cost per success crosses the next model's after one or two failures; fixed plans collapse at different loads; late passes are confirmed less often by a single judge.}
\label{fig:motiv}
\end{figure}

\begin{figure*}[t]
\centering
\includegraphics[width=\textwidth]{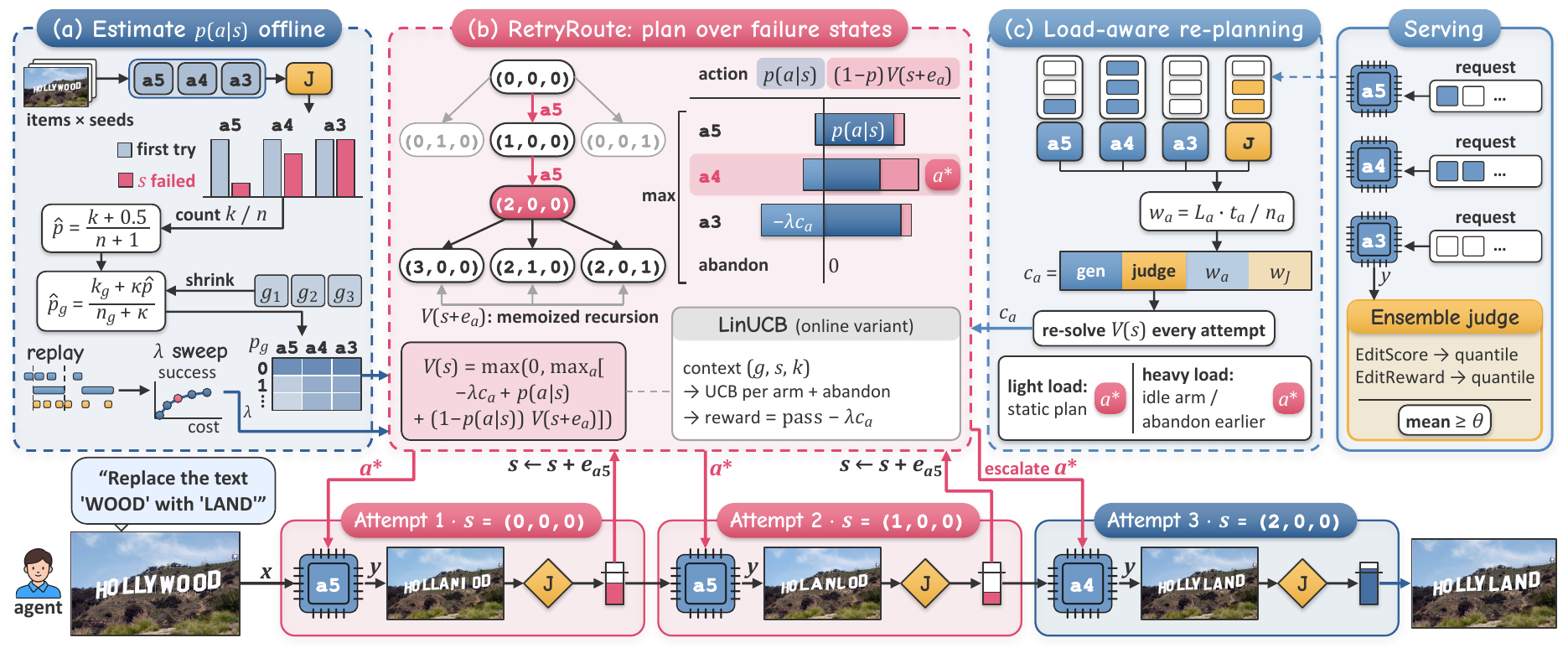}
\caption{Overview of \sys. It (a) estimates $p(a\mid s)$ offline, (b) plans over failure states and (c) re-plans with queue waits. In the bottom row a5 fails twice and \sys escalates to a4.}
\Description{System diagram: offline estimation of p(a|s) from logged attempts, the RetryRoute planner over failure states with a Bellman equation and LinUCB variant, load-aware re-planning from queue waits, GPU workers with an ensemble judge, and the running example of three attempts at the bottom.}
\label{fig:overview}
\end{figure*}

\subsection{The Agentic Editing Loop}
\label{sec:formulation}

\paragraph{Agentic editing loop.}
A request $x$, an image with an instruction, is served by a sequence of attempts.
Each attempt picks an editing model, an \emph{arm} $a\in\mathcal{A}$ tried at most $m_a$ times, pays cost $c_a$ for generation plus one 10\,s judge call, and is accepted or rejected by a judge. The request ends at the first acceptance or when the policy abandons it.
Our two pools of open models span 30$\times$ and 190$\times$ in generation time on 24\,GB RTX 4090D GPUs (Table~\ref{tab:arms}).

\begin{table}[t]
\caption{Editing models, called arms, and GPU seconds per edit, with pool~1 above the rule and pool~2 below.}
\label{tab:arms}
\small
\setlength{\tabcolsep}{3.5pt}
\begin{tabular*}{\linewidth}{@{\extracolsep{\fill}}llrr}
\toprule
Arm & Model & Steps & Time (s) \\
\midrule
a5 & FLUX.2 [klein] 4B~\cite{flux2klein} & 4 & 5.7 \\
a4 & Qwen-Image-Edit-2511~\cite{qwenimage} + Lightning~\cite{lightning} & 8 & 39.8 \\
a3 & Qwen-Image-Edit-2511 & 40 & 170.7 \\
a1 & FLUX.1 Kontext-dev~\cite{kontext} & 28 & 60.2 \\
a2 & a1 + Turbo-Alpha~\cite{turboalpha} & 8 & 25.7 \\
\midrule
b1 & VIBE~\cite{vibe} & 20 & 0.9 \\
b2 & LongCat-Image-Edit~\cite{longcat}, CPU offload & 50 & 108 \\
b3 & Step1X-Edit v1.2~\cite{step1xedit,reasonedit}, 2 GPUs & 28 & 173 \\
\bottomrule
\end{tabular*}
\end{table}

\paragraph{Planning over failure states.}
Because a request continues only after a rejection, every policy is a sequence of attempts with value $\sum_j R_j\,(p_j-\lambda c_j)$, where $p_j$ is the success probability of the $j$-th attempt given that all earlier ones failed, $c_j$ its GPU seconds, $R_j=\prod_{i<j}(1-p_i)$ and $\lambda$ a cost weight.
With the failures so far as the state $s=(f_a)_{a\in\mathcal{A}}$, let $p(a\mid s)$ be the probability that the next attempt on $a$ is accepted after the rejections in $s$. The optimal policy then solves the Bellman equation~\cite{bellman}:
\begin{equation}
V(s)=\max\Big(0,\ \max_{a:\,f_a<m_a}\big[-\lambda c_a + p(a\mid s) + (1-p(a\mid s))\,V(s+e_a)\big]\Big),
\label{eq:bellman}
\end{equation}
where $0$ is abandonment and $e_a$ the unit vector of $a$, and sweeping $\lambda$ traces a cost--success frontier.
Same-model retry, per-request routers and cascades all plan with the first-try rate $p(a\mid\mathbf{0})$ at the state $\mathbf{0}$ with no failures in place of $p(a\mid s)$, which is exact only if retries are independent.

\subsection{Failures Are Sticky}
\label{sec:measure}

\paragraph{Benchmarks and judges.}
We use 438 GEdit-Bench~\cite{step1xedit} items in 11 edit categories with Chinese and with English instructions, 230 items of GEditBench v2~\cite{geditv2}, and 180 ImgEdit~\cite{imgedit} items from 9 categories.
Each model is run with 8 seeds for the cheapest arm and 2 seeds for the others, and the $k$-th attempt of a model on an item is its seed $k{-}1$.
Two independent judges score every image.
EditScore~\cite{editscore} on Qwen3-VL-8B~\cite{qwen3vl} uses its pass threshold $\tau=7.5$, and EditReward~\cite{editreward} on Qwen2.5-VL-7B~\cite{qwen25vl} uses a threshold set to give the same overall pass rate.
In the loop we use an ensemble of the two vision-language judges, which averages each score's quantile over all images and passes an attempt above a threshold $\theta$ that gives the pass rate of EditScore at $\tau=7.5$.
Pool~1 on Chinese and English GEdit-Bench (GEdit-CN and GEdit-EN), a five-arm English set, GEditBench v2 and ImgEdit, and pool~2 on Chinese and English GEdit-Bench and ImgEdit form eight groups, which together hold 30{,}042 edited images.

\paragraph{A retry rarely succeeds, but another model often does.}
In all 50 model--judge cells, a retry succeeds only 17--53\% as often as a first try, and Figure~\ref{fig:sticky}a averages them per model.
For a5 on English GEdit-Bench the next a5 attempt passes 0.143 after one failure, while a4 still passes 0.436 (Figure~\ref{fig:motiv}a), and eight a5 attempts reach 0.804 where independence predicts 1.000.
\textbf{We call this \sticky.}
\emph{The next attempt's chance depends on the failure history}, which motivates estimating $p(a\mid s)$ per failure state (\S\ref{sec:estimate}).

\paragraph{Latent difficulty explains it.}
Sticky failures are what heterogeneous request difficulty predicts, as Appendix~\ref{app:proofs} proves.
\begin{lemma}
Let a request have a latent difficulty $z$, and let attempts of arm $a$ on it succeed independently with probability $q_a(z)$. Then a history of rejections affects the next attempt only through the failure counts $s$, so (\ref{eq:bellman}) over $s$ is exact. With $\mathbb{E}_s$ the expectation given $s$:
\begin{align*}
p(b\mid s{+}e_a)&=p(b\mid s)-\mathrm{Cov}_s(q_a,q_b)\,/\,\mathbb{E}_s[1-q_a],\\
p(a\mid s{+}e_a)&=p(a\mid s)-\mathrm{Var}_s(q_a)\,/\,\mathbb{E}_s[1-q_a].
\end{align*}
\end{lemma}
A retry is therefore never more likely to succeed than the attempt it follows, and strictly less likely when requests differ in difficulty for that model.
With shared difficulty, where $\mathrm{Cov}>0$, a failure also lowers the other models' chances, as a4's drop from 0.733 to 0.436 after one a5 failure shows.

\paragraph{When to escalate depends on the failures so far.}
Per success, a second a5 attempt costs as much as an a4 attempt (110\,s against 114\,s) and a third twice as much (Figure~\ref{fig:motiv}b).
Choosing the model up front adds only about one point, since a5 is the cheapest per success in all 11 categories.
\emph{The decision that matters comes after each failure}, which motivates planning over failure states (\S\ref{sec:act}).

\paragraph{Load flips the best fixed plan.}
The fixed plan a5$\times$6$\to$a4 is best at 0.02\,req/s with goodput 82.7\% but drops to 9.3\% at 0.06, where a5$\to$a4 keeps 74.2\% (Figure~\ref{fig:motiv}c).
Goodput is the share of requests that succeed within the service-level objective (\S\ref{sec:eval}).
\emph{No fixed plan is best at every load}, which motivates re-planning with live queue waits after every attempt (\S\ref{sec:act}).

\paragraph{Late passes are less reliable.}
EditReward confirms 84--93\% of the in-loop judge's first-try passes but only 51--75\% of its late passes, those after a failure, and this rate is lower in all eight groups (Figure~\ref{fig:motiv}d).
\emph{Many attempts on a weak model collect false passes}, so a retry loop should trust a late pass less (\S\ref{sec:conclusion}).

\section{RetryRoute Design}
\label{sec:design}

\sys rests on one quantity, $p(a\mid s)$, which an estimator learns from logged attempts for C1 and a planner turns into the next action after every rejection for C2 (Figure~\ref{fig:overview}).
\S\ref{sec:why} then shows when simpler fixed rules are already optimal.

\subsection{Estimating Success Given Failures}
\label{sec:estimate}
To address C1, \sys estimates $p(a\mid s)$ by counting.
Among logged training requests whose attempts in $s$ all failed, it takes the fraction whose next attempt on $a$ passed.
States far down the tree have few requests, so we estimate per request category and shrink toward the all-category estimate~\cite{efronmorris}:
\begin{equation}
\hat p_g(a\mid s)=\frac{k_g+\kappa\,\hat p(a\mid s)}{n_g+\kappa},\qquad \hat p(a\mid s)=\frac{k+0.5}{n+1},
\end{equation}
where $k_g$ of the $n_g$ category-$g$ requests in state $s$ passed on $a$, $k$ of $n$ over all categories, and $\kappa=20$.
This keeps the sharp drop after a failure, which first-try rates $p(a\mid\mathbf{0})$ miss, while a category with few logs falls back to the pooled estimate.
\paragraph{Counting needs no model of difficulty.} We count rather than fit the latent-difficulty model of the Lemma, since counts need no parametric form for $q_a(z)$ and are most accurate in the shallow states that most requests visit.

\subsection{Planning Retries under Load}
\label{sec:act}
\paragraph{Exact planning.}
To address C2, \sys acts on the estimates by exact planning or, for cold start and drift, by online learning, and it adds the load to the cost.
The state space is small, with $\prod_a(m_a+1)$ states, for example 81 for three arms with caps 8, 2 and 2.
Memoized recursion solves (\ref{eq:bellman}) exactly in 0.35\,ms for three arms and 5.6\,ms for five, and we call it the dynamic-programming (\emph{DP}) planner.

\paragraph{Load-aware re-planning.}
In a shared service, an attempt on a busy arm also costs queueing delay.
At each decision, \sys adds to $c_a$ the current expected wait of arm $a$'s queue and of the judge queue. Each wait follows Little's law~\cite{little} as queue length times measured service time divided by the number of workers, and \sys then re-solves (\ref{eq:bellman}) from the request's current state.
Under light load this reduces to the static plan, and under heavy load it shifts attempts toward idle arms or abandons earlier.

\paragraph{Online learning.}
For cold start or drift, \sys can instead learn online with LinUCB~\cite{auer,linucb} plus an abandon action.
The context at each attempt is the request category, the number of failures so far on each arm and the attempt index, and the reward of an attempt is its acceptance minus $\lambda c_a$.
Without the failure history it treats every retry as a first attempt, as ACQB~\cite{acqb} does.

\subsection{When Fixed Rules Are Optimal}
\label{sec:why}
Index rules are optimal for independent alternatives in classical index and search theory~\cite{gittins,weitzman,simonkadane}, and here fixed rules are optimal only in special cases, with proofs in Appendix~\ref{app:proofs}.
\begin{proposition}
(i) If retries are independent, $p(a\mid s)=p_a$, an optimal policy tries the arms in increasing order of cost per success $c_a/p_a$, each up to $m_a$ times, skipping arms with $\lambda c_a/p_a\geq 1$. With $m_a=\infty$ it repeats the single arm $\arg\min_a c_a/p_a$.
(ii) If a failure changes only the failed arm's own success rate, $p(a\mid s)=p_a(f_a)$ non-increasing in $f_a$, the index rule is optimal. It tries the arm with the smallest current $c_a/p_a(f_a)$, stops when $\lambda c_a/p_a(f_a)\geq 1$, and in particular escalates from $a$ to $b$ as soon as $c_a/p_a(f_a)>c_b/p_b(f_b)$.
\end{proposition}
Part (i) is the logic behind retry loops on a cheap model, and part (ii) says when to escalate.
For a5 the switch comes after the first or second failure (Figure~\ref{fig:motiv}b).
With shared difficulty, the Cov term of the Lemma, no fixed index is optimal in general.
The next arm's value depends on which arms have failed, and only (\ref{eq:bellman}) over the failure counts captures it.
For example, let 70\% of requests be easy and the rest hard for both arms (Appendix~\ref{app:proofs}).
The index rule then tries a, a, b and is worth 0.514, while the planner over failure counts is worth 0.560.

\section{Evaluation}
\label{sec:eval}

\input{tables/tab_main}

\subsection{Setup}
\paragraph{Protocol.}
Offline results replay the logged edits of \S\ref{sec:measure}, so every success is a real model output accepted by the judges and every cost is a measured GPU time from Table~\ref{tab:arms}.
Every method tunes its own knobs on the training folds of a 5-fold split and is reported on the test folds. At a budget $B$, the mean GPU seconds per request with judge calls, each method contributes its best point whose test cost is at most $B$.
The main budget is 50\,s for pool~1 and 120\,s for pool~2, about one a4 or LongCat attempt, and differences are paired per item with bootstrap~\cite{efron79} 95\% confidence intervals (CIs).
Two numbers summarize a whole curve over a range of budgets (Appendix~\ref{app:setup}).
The average performance gap recovered (APGR) of RouteLLM~\cite{routellm} is the mean share of the gap between one attempt of the cheapest model and the strongest single model that a method closes, and the area under the curve (AUC) is the mean success.
\paragraph{Baselines.}
Same-model retry and Best-of-N each retry one arm. Rewrite-retry, after the critique-and-revise pattern~\cite{selfrefine,idea2img}, has a vision-language model rewrite the instruction before each retry of the cheap arm, from the source alone (blind) or also from the failed edit and its score (critique).
The routers are Cost-Aware Routing~\cite{costaware} (CA-Routing) and the similarity-weighted (SW) and matrix-factorization (MF) routers of RouteLLM~\cite{routellm}, and the cascades are FrugalGPT~\cite{frugalgpt}, AutoMix~\cite{automix}, DiffServe~\cite{diffserve} and HADIS~\cite{hadis}.
Ours are the DP planner, LinUCB, a fixed ladder of arms and a queue threshold (Appendix~\ref{app:setup}).
The controls are the DP fed first-try rates and LinUCB without failure history, which resembles ACQB, and the oracle picks the cheapest successful attempt per item.
Table~\ref{tab:main} shows the oracle and the controls in gray rows and marks the top two in bold and underline.
\paragraph{Serving testbed.}
\sys runs on three RTX 4090D GPUs, one each for a5, a4 and the judge, and serves 160 GEdit-Bench requests at two steady loads and a bursty one (Appendix~\ref{app:testbed}).
A request counts toward \emph{goodput}, which we adapt from DistServe~\cite{distserve}, if it succeeds within a 300\,s service-level objective. The DP uses one setting ($\lambda=0.0018$) at all loads.

\subsection{Main Results}
\label{sec:offline}

\begin{figure}[t]
\centering
\includegraphics[width=\linewidth]{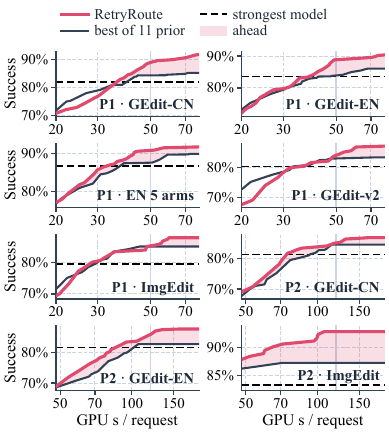}
\caption{\sys against the best of eleven prior baselines at each budget, with the main budget as a vertical line.}
\Description{Eight panels in four rows and two columns, one per group, five of pool 1 (P1) and three of pool 2 (P2), of success rate against GPU seconds per request; the RetryRoute curve lies on or above the envelope of the prior baselines at most budgets.}
\label{fig:curves}
\end{figure}

\begin{figure}[t]
\centering
\includegraphics[width=\linewidth]{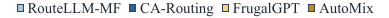}\\
\begin{subfigure}[t]{0.989\linewidth}\centering\includegraphics[width=\linewidth]{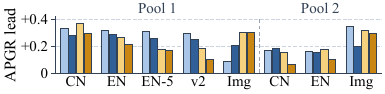}\end{subfigure}\\[1pt]
\begin{subfigure}[t]{0.989\linewidth}\centering\includegraphics[width=\linewidth]{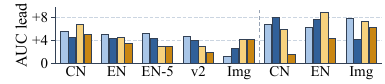}\end{subfigure}
\caption{Lead of \sys in APGR and AUC over four prior baselines in each group.}
\Description{Two bar panels over the eight groups, five groups of pool 1 and three of pool 2, with four bars per group for the lead of RetryRoute over RouteLLM-MF, CA-Routing, FrugalGPT and AutoMix. Every bar is above zero, from about 0.07 to 0.37 in APGR and from about 1.2 to 8.9 points in AUC.}
\label{fig:cvsum}
\end{figure}

\begin{figure}[t]
\centering
\includegraphics[width=\linewidth]{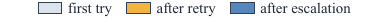}\\
\includegraphics[width=\linewidth]{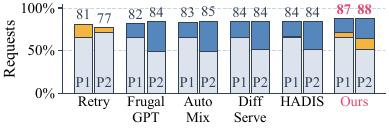}
\caption{How requests end at the main budget for pool~1 and pool~2, with success in percent on top of each bar.}
\Description{Stacked bars per method and pool: share of requests passing on the first try, after a same-model retry, or after escalation.}
\label{fig:usage}
\end{figure}

\begin{figure}[t]
\centering
\includegraphics[width=\linewidth]{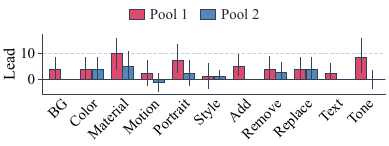}
\caption{Lead of \sys over the strongest prior baseline in each GEdit-Bench edit category, pooling Chinese and English, with 95\% CIs.}
\Description{Grouped bars per edit category for the two pools; the lead is positive in most categories.}
\label{fig:category}
\end{figure}

\begin{figure*}[t]
\centering
\begin{minipage}[b]{0.775\textwidth}\includegraphics[width=\linewidth]{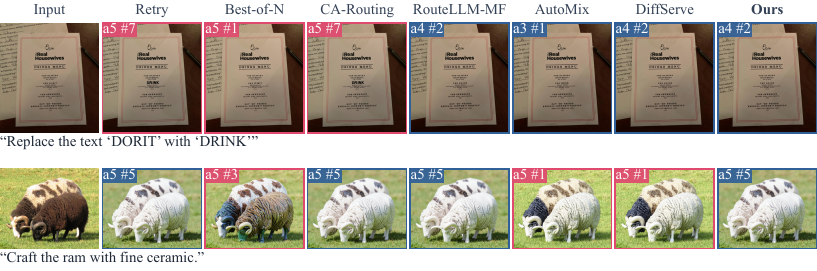}\end{minipage}\hfill
\begin{minipage}[b]{0.205\textwidth}
\captionsetup{justification=raggedright,singlelinecheck=false}
\caption{What each method delivers at the main budget on two pool-1 GEdit-Bench requests in English, framed blue for a pass and pink for a failure, with the arm and attempt on each badge.}
\Description{Rows of requests; for each, the input and the image delivered by six prior baselines and RetryRoute, framed blue for a pass and pink for a failure.}
\label{fig:ex-en}
\end{minipage}
\end{figure*}

\paragraph{Conditioning on failures beats prior routing.}
At the main budget \sys is never below any of the eleven prior baselines (Table~\ref{tab:main}).
The methods that retry one model trail it the most in every group, and its lead over the strongest prior baseline is significant in five.
The DP also beats the strongest single model in every group, at a lower cost in seven.
Fed first-try rates instead, the same planner drops significantly on four of the five pool-1 groups, so there the gain comes from conditioning on failures.
The cheapest setting of OctoT2I~\cite{octot2i} already exceeds the main budget in five groups, so Table~\ref{tab:octo} compares the two at OctoT2I's own cost for round limits of one to four.
There \sys is significantly more successful in 27 of the 32 settings and never significantly less, with a lead of up to 17.35 points at a round limit of one.

\paragraph{The lead holds over most budgets.}
Even against the best prior baseline picked separately at every budget, \sys is ahead over most of the range (Figure~\ref{fig:curves}).
It matches the strongest single model at a lower cost in every group, up to 6.4$\times$ lower.

\paragraph{\sys leads on both curve summaries.}
In every group \sys has a higher APGR and AUC than every prior baseline, and Figure~\ref{fig:cvsum} draws its lead over the four that are strongest in some group.
In six of the eight groups its APGR exceeds one, so over the budget range it gains more than the whole gap between the cheapest and the strongest single model.

\paragraph{The gain comes from both retrying and escalating.}
The cascades almost never retry and same-model retry never escalates.
\sys retries the cheap arm while a retry is still cheaper per success, escalates otherwise and fails on the fewest requests (Figure~\ref{fig:usage}).
Figure~\ref{fig:ex-en} shows both on two pool-1 requests in English.
To replace a printed word, \sys escalates to a4 after one failure on a5, while the three methods that stay on a5 fail.
To make the ram ceramic, it passes on its fifth a5 attempt, while Best-of-N, AutoMix and DiffServe deliver an earlier a5 attempt that fails.
The gain is also broad across edit types, positive in 17 of the 22 category and pool pairs (Figure~\ref{fig:category}).

\paragraph{Rewriting helps but does not replace escalation.}
Rewriting the instruction before a retry is a strong low-cost baseline.
At the main budget \sys significantly beats both rewrite variants in 14 of the 16 settings, while near the rewrite loop's own spend the two mostly tie, so the gain comes from spending the rest of the budget on escalation.
Rewriting is also complementary, since as one more action inside \sys it never lowers success significantly and adds at most 2.2 points.

\subsection{Ablations and Robustness}
\label{sec:ablation}

\begin{figure}[t]
\centering
\includegraphics[width=\linewidth]{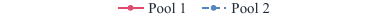}\\
\begin{subfigure}[t]{0.502\linewidth}\centering\includegraphics[width=\linewidth]{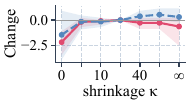}\caption{Shrinkage $\kappa$, default 20.}\end{subfigure}\hfill
\begin{subfigure}[t]{0.487\linewidth}\centering\includegraphics[width=\linewidth]{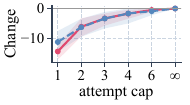}\caption{Cap on attempts per request.}\end{subfigure}\\[1pt]
\begin{subfigure}[t]{0.502\linewidth}\centering\includegraphics[width=\linewidth]{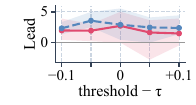}\caption{Judge threshold shift.}\end{subfigure}\hfill
\begin{subfigure}[t]{0.487\linewidth}\centering\includegraphics[width=\linewidth]{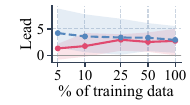}\caption{Training data for all methods.}\end{subfigure}
\caption{Sensitivity in points. Panels a and b give the change from the default \sys, and panels c and d the lead over the best prior baseline.}
\Description{Four panels: shrinkage, attempt cap, judge threshold and training-data fraction.}
\label{fig:sens}
\end{figure}

\paragraph{Multi-attempt planning matters most.}
Capping a request at two attempts loses significantly in every group (Figure~\ref{fig:sens}b), and allowing the cheap arm only once loses significantly in six of the eight.
Every moderate shrinkage $\kappa$ stays close to the default, while no shrinkage at $\kappa=0$ loses significantly in four of the eight groups (Figure~\ref{fig:sens}a).
In the online learner the failure history helps on pool~2 (Table~\ref{tab:main}).

\paragraph{The lead is robust to the judge and the training data.}
Shifting the pass threshold in either direction keeps \sys at or above the best prior baseline in 30 of the 32 shifted cells (Figure~\ref{fig:sens}c).
With less training data, given to every method, its lead stays positive on average in both pools down to 5\% of the logs (Figure~\ref{fig:sens}d).
Flipped training verdicts never put \sys significantly behind at any tested rate, and a coarser cost-weight grid moves the lead by less than half a point in most settings.
Over six random splits into folds, \sys is never below the strongest of the nine prior baselines that do not rewrite and never significantly below any of them.

\subsection{Generalization and Online Learning}

\input{tables/tab_xfer}

\begin{figure}[t]
\centering
\includegraphics[width=\linewidth]{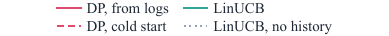}\\
\begin{subfigure}[t]{0.499\linewidth}\centering\includegraphics[width=\linewidth]{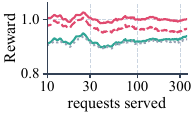}\caption{Cold start, relative reward.}\end{subfigure}\hfill
\begin{subfigure}[t]{0.490\linewidth}\centering\includegraphics[width=\linewidth]{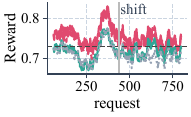}\caption{Drift, EN $\to$ ImgEdit.}\end{subfigure}
\caption{Online learning from a cold start and after a shift.}
\Description{Learning curves of the offline and online planners and of LinUCB with and without retry history.}
\label{fig:online}
\end{figure}

\paragraph{\sys transfers to unseen categories and benchmarks.}
With each category held out from training in turn, the DP barely changes and stays above the best prior baseline in all eight groups (Table~\ref{tab:xfer}).
Trained on another benchmark of the same pool, it is at least as high as the best prior baseline transferred the same way in every train and test pair, and the table averages two sources per test set.
Bold in Table~\ref{tab:xfer} marks where Ours is at least as high as Prior.

\paragraph{\sys carries over to text-to-image generation.}
We run the prompts of GenEval~\cite{geneval} with the text-to-image tools of OctoT2I~\cite{octot2i} and its NVILA verifier~\cite{nvila} in the loop (Appendix~\ref{app:more}).
At the main budget \sys passes the verifier on 91.9\% of prompts, the oracle rate, and is significantly above 14 of the 15 compared methods.
At equal cost it is also significantly above OctoT2I at every round limit whose cost \sys can match, two to four rounds.

\paragraph{\sys can start without logs.}
Starting cold and learning only from its own attempts, the DP reaches 97\% of the reward of the DP trained on full logs within 100 requests, and the failure history also helps LinUCB (Figure~\ref{fig:online}a).
Under a shift from English GEdit-Bench to ImgEdit, both the planner trained before the shift and the online learners hold their reward (Figure~\ref{fig:online}b).

\subsection{Planning Cost and Number of Models}
\label{sec:scale}

\paragraph{Planning is cheap.}
One exact solve takes well under a millisecond for up to three arms, eight arms still plan in 0.2\,s, and plans are cached per category and load level.
A LinUCB decision takes 92\,$\mu$s, and on the testbed the router adds a median 6.9\,ms per decision against attempts that take seconds.

\paragraph{More models keep the lead.}
The lead grows from two arms to three and then stays flat, and with all six models of both pools in one loop \sys leads the strongest prior baseline on all three test sets at both budgets, significantly in five of the six cases.
On a third pool, FLUX.2 [klein] 9B~\cite{flux2klein9b}, Qwen-Image-2.1~\cite{qwenimage21} and GLM-Image~\cite{glmimage}, \sys passes 91.8\% on GEdit-EN at the main budget, the oracle rate.
It is significantly above all nine prior baselines that do not rewrite the instruction (Appendix~\ref{app:more}).

\subsection{Serving Testbed}
\label{sec:serving}

\input{tables/tab_serve}

\paragraph{Load awareness pays off on GPUs.}
The DP has the highest goodput at all three loads, stays within the service-level objective, and is significantly above the five policies without load awareness in 10 of the 15 comparisons by a two-proportion test (Table~\ref{tab:serve}).
HADIS and the fixed ladder pile attempts onto the single a4 GPU and miss the service-level objective.
Offline, without queues, the DP and the fixed ladder are within about a point, so the gap comes mostly from load awareness.
At 0.055\,req/s the DP waits a third as long in queues as the fixed ladder.
A second testbed serves pool~2 on three GPUs at one steady load, where \sys reaches a goodput of 81.2\% against 67.5\% for same-model retry, the best of the four prior methods run there (Appendix~\ref{app:testbed}).

\begin{figure}[t]
\centering
\includegraphics[width=\linewidth]{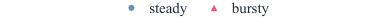}\\
\begin{subfigure}[t]{0.359\linewidth}\centering\includegraphics[width=\linewidth]{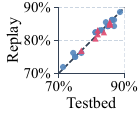}\caption{Success.}\end{subfigure}\hfill
\begin{subfigure}[t]{0.301\linewidth}\centering\includegraphics[width=\linewidth]{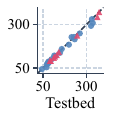}\caption{Mean latency.}\end{subfigure}\hfill
\begin{subfigure}[t]{0.320\linewidth}\centering\includegraphics[width=\linewidth]{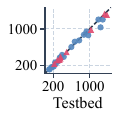}\caption{p95 latency.}\end{subfigure}
\caption{Load replay against testbed over 30 runs on the same request streams, with latency in seconds on log axes.}
\Description{Scatter plots of replayed against measured success and latency, close to the equality line.}
\label{fig:calib}
\end{figure}

\paragraph{Load replay matches the testbed.}
For loads the testbed does not cover, \emph{load replay} runs the router's planner on the logged edits, with their verdicts and measured service times, in a discrete-event model of the GPU queues (Appendix~\ref{app:testbed}).
Over 30 testbed runs on two machines, replayed success is within 2.5 points of the measured one and the median error of mean latency is about 1\% (Figure~\ref{fig:calib}).

\paragraph{A heuristic tuned per load is not enough.}
Table~\ref{tab:qth} tunes the queue threshold and the ladder at one of three loads and runs each in load replay at steady loads from 0.04 to 0.06\,req/s and a bursty one.
The threshold nears the DP only when tuned for the right load, and one picked at 0.045\,req/s loses over a third of its goodput at 0.06, while one $\lambda$ keeps the DP best at every load.

\begin{figure}[t]
\centering
\includegraphics[width=\linewidth]{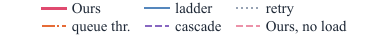}\\
\begin{subfigure}[t]{0.519\linewidth}\centering\includegraphics[width=\linewidth]{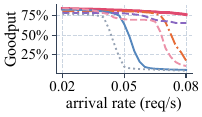}\caption{Load sweep, 3 GPUs.}\end{subfigure}\hfill
\begin{subfigure}[t]{0.470\linewidth}\centering\includegraphics[width=\linewidth]{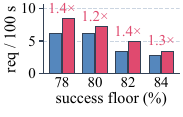}\caption{Capacity, 3 GPUs.}\end{subfigure}
\caption{Goodput in load replay and the highest load sustained at each success floor within the service-level objective.}
\Description{Left: goodput curves over arrival rate. Right: bars of the highest sustainable arrival rate of RetryRoute and the best fixed ladder at four success floors.}
\label{fig:load}
\end{figure}

\paragraph{\sys sustains more load than fixed plans.}
In load replay the DP keeps the highest goodput from 0.03\,req/s upward, while the fixed ladder and later the threshold tuned at 0.055 collapse (Figure~\ref{fig:load}a).
At the same success floor and service-level objective it sustains up to 1.4$\times$ the arrival rate of the best fixed ladder on three GPUs, the ratio written above the bars of Figure~\ref{fig:load}b.

\begin{figure}[t]
\centering
\includegraphics[width=\linewidth]{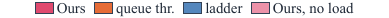}\\
\begin{subfigure}[t]{0.518\linewidth}\centering\includegraphics[width=\linewidth]{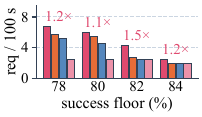}\caption{Bursty stream.}\end{subfigure}\hfill
\begin{subfigure}[t]{0.471\linewidth}\centering\includegraphics[width=\linewidth]{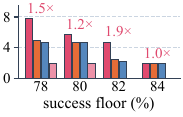}\caption{DiffusionDB trace.}\end{subfigure}
\caption{Highest load sustained in load replay at each success floor within the service-level objective on three GPUs, under bursty arrivals and an arrival trace.}
\Description{Two bar panels over success floors of 78, 80, 82 and 84 percent. In each group the RetryRoute bar is the tallest or tied, above the best queue threshold, the best fixed ladder and the DP without load awareness, with ratios over the better heuristic from 1.0 to 1.9 written above the groups.}
\label{fig:burst}
\end{figure}

\paragraph{The gain persists under bursty arrivals.}
With the bursty stream of the testbed and with a DiffusionDB arrival trace, \sys sustains up to 1.5$\times$ and 1.9$\times$ the arrival rate of the stronger tuned heuristic, ladder or queue threshold, respectively. Figure~\ref{fig:burst} writes this ratio above the bars at each floor.

\subsection{Judge Robustness}
\label{sec:judge}

\paragraph{Either single judge in the loop keeps the lead.}
With EditScore or EditReward in the loop instead of the ensemble, Table~\ref{tab:loopjudge} gives the success of \sys and of the strongest prior baseline in each group.
\sys is never significantly below any of the eleven prior baselines in the eight groups and is significantly above in 83 and 75 of the 88 comparisons, respectively.

\paragraph{Annotators confirm the in-loop judge.}
Two members of our research group labeled a sample of attempts stratified by group and by the judges' verdicts, half of them near the pass line. They read each English instruction with a machine translation as an aid, and we reweight the sample to all attempts with equal weight per group.
Against their majority vote the in-loop ensemble is right on 83.7\% of attempts, significantly more than either single judge, with only 2.6\% false accepts (Table~\ref{tab:human}).
The annotators agree with a Cohen's kappa of 0.86.

\paragraph{At the main budget a quality gate does not put \sys behind.}
On pool~2 we also gate each pass on a VIEScore~\cite{viescore} perceptual-quality bar, set at three levels around LongCat's median first-try score, and refit every method (Appendix~\ref{app:more}).
\sys is never significantly below the strongest prior baseline and is significantly above a single LongCat try in 7 of the 9 settings. Averaged over the three sets, its margin shrinks as the bar rises until it ties the strongest prior baseline at the strictest bar.
With the same gate on pool~1, \sys ties the strongest prior baseline at the middle bar in every set and leads it significantly in three sets at the lowest bar of the three.

\section{Conclusion}
\label{sec:conclusion}
\label{sec:concl}
We presented \sys, a retry-aware router for agentic image editing services, motivated by \sticky.
A model that has failed an edit will most likely fail it again, while a different model often still succeeds.
\sys estimates each model's chance of success given the request's failure history and plans retries, escalations and abandonment over failure states, with live queue waits in the cost.
At the main budget and under the in-loop judge it leads the strongest of eleven prior baselines by 2.8 points on average over eight offline groups.
On a GPU serving testbed its goodput exceeds the best prior baseline's by 5.0--6.3 points at every load with one setting.
The correlation also reaches the judges, so retry loops should be stricter about late passes.
We expect failure-aware planning to help other tools in agent loops, such as code interpreters and search engines that agents call again after a rejected answer.

\bibliographystyle{ACM-Reference-Format}
\bibliography{refs_gen,refs_manual}

\appendix
\section*{Appendix}

\section{Proofs}
\label{app:proofs}

\paragraph{Policies are sequences.}
A request continues only after a rejection, so before its $j$-th attempt a policy always knows the same thing, namely that all earlier attempts failed.
A deterministic policy is therefore a fixed sequence of arms with a stopping point, and its value is $U=\sum_j R_j(p_j-\lambda c_j)$ with $R_j=\prod_{i<j}(1-p_i)$.

\paragraph{Lemma.}
Given $z$, attempts are independent, so a rejection history with failure counts $s$ has likelihood $\prod_a(1-q_a(z))^{f_a}$, which depends on the history only through $s$.
The posterior is $\pi(z\mid s)\propto\pi(z)\prod_a(1-q_a(z))^{f_a}$ and the next attempt on $b$ succeeds with $p(b\mid s)=\mathbb{E}_s[q_b]$, hence (\ref{eq:bellman}) over $s$ is exact.
One more failure on $a$ reweights the posterior by $(1-q_a)/\mathbb{E}_s[1-q_a]$, so $p(b\mid s{+}e_a)=\mathbb{E}_s[q_b(1-q_a)]/\mathbb{E}_s[1-q_a]=\big(p(b\mid s)(1-p(a\mid s))-\mathrm{Cov}_s(q_a,q_b)\big)/\mathbb{E}_s[1-q_a]$, which is the stated identity.
For $b=a$ the covariance is a variance and the change is never positive. $\square$

\paragraph{Proposition.}
Take two adjacent attempts $i,i{+}1$ on different arms, and assume a failure changes only the failed arm's own rate.
Swapping them leaves $R_{i+2}=R_i(1-p_i)(1-p_{i+1})$ and every later rate unchanged, and $U(i\text{ first})-U(i{+}1\text{ first})=\lambda R_i\,p_ip_{i+1}\,(c_{i+1}/p_{i+1}-c_i/p_i)$.
So an optimal sequence can be taken in non-decreasing order of $c/p$~\cite{simonkadane}.
Since $p_a(f)$ does not increase with $f$, each arm's own ratios increase along its attempts, and the sorted order is the one the index rule produces.
An attempt adds $R_j(p_j-\lambda c_j)$, which is positive exactly when $\lambda c_j/p_j<1$.
A sorted tail whose ratios are all at least $r$ is worth at most $1-\lambda r$ per unit of reach, so inserting any attempt with $\lambda c/p<1$ at its sorted position never lowers $U$, and the optimal sequence is the sorted set of all such attempts.
Part (i) is the case $p_a(f)=p_a$, and with $m_a=\infty$ the sequence repeats $\arg\min_a c_a/p_a$ and is worth $1-\lambda c_a/p_a$. $\square$

\paragraph{Shared difficulty breaks the index rule.}
The example of \S\ref{sec:why} lets 70\% of requests be easy with $q_a=0.9$ and $q_b=0.95$, and 30\% hard with $q_a=0.05$ and $q_b=0.30$, with $c_a=10$, $c_b=40$, $m_a=2$, $m_b=1$ and $\lambda=0.012$.
The index rule there uses each arm's own failure counts.
We also checked the Lemma and both parts of the Proposition numerically.
On 2{,}000 random latent-difficulty instances the identity holds to $10^{-15}$, and on 300 random instances each the index rule and the DP match an exhaustive search.

\section{Model--Judge Cells}
\label{app:cells}
Table~\ref{tab:cells} summarizes per group the first-try and retry pass rates of the 50 cells behind Figure~\ref{fig:sticky}a, thresholding EditReward at the pass rate of EditScore at $\tau=7.5$.

\input{tables/tab_cells}

\section{Full Cost--Success Curves}
\label{app:curves}
\begin{figure*}[t]
\centering
\includegraphics[width=\textwidth]{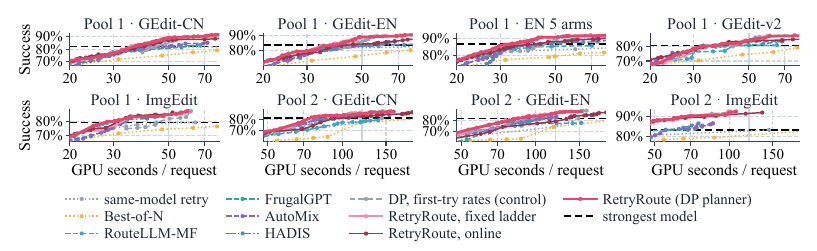}
\Description{Eight panels with one cost--success frontier per method.}
\caption{Cost--success frontiers of each method.}
\label{fig:curvesall}
\end{figure*}

Figure~\ref{fig:curvesall} shows the test-fold Pareto frontier of each method, leaving out Cost-Aware Routing, RouteLLM-SW and DiffServe for legibility.

\section{Example Selection}
\label{app:examples}

The rows of Figure~\ref{fig:ex-en} come from distinct edit categories, and each is a request on which \sys passes and at least two of the six shown baselines fail, screened for content suitable for display and not used in Figure~\ref{fig:sticky}.
The second row is chosen among such requests on which DiffServe also fails, and the other rows are drawn at random with a fixed seed.
Frames show the in-loop verdict, blue for a pass and pink for a failure, and each badge names the arm and attempt that produced the image.

\section{Protocol and Baselines}
\label{app:setup}
Offline results replay logged edits, as the offline evaluations of RouteLLM, FrugalGPT and AutoMix replay logged outputs~\cite{routellm,frugalgpt,automix}.
On pool-1 ImgEdit, offline replays use only arms a5 and a4.
APGR averages over budgets from 20 to 100\,s for pool~1 and from 10 to 220\,s for pool~2.
AUC is the mean success over budgets from 30 to 80\,s for pool~1 and from 40 to 160\,s for pool~2.
The ladder is a fixed plan as in \S\ref{sec:measure}, the single arm sequence, such as a5$\times$3$\to$a4, that scores best on the training folds, and the queue threshold falls back to the cheap arm when the next arm's wait is too long.
Cost weights lie on a common grid of 97 values from 0.0003 to 0.03, CIs use 2{,}000 bootstrap resamples, and unless stated otherwise every experiment uses the in-loop judge, folds, grid and budget rule of \S\ref{sec:eval}.
Same-model retry picks the best arm and $k$.
The rewriter of rewrite-retry is Qwen3-VL-8B~\cite{qwen3vl}, charged 2\,s per rewrite, and the judge keeps the original instruction.
CA-Routing picks one arm per category and retries it up to $k$ times, and the RouteLLM routers read instruction and source-image embeddings.
FrugalGPT learns its order, and AutoMix runs a partially observable Markov decision process (POMDP) over judge-score buckets.
The queue threshold follows a ladder such as a5$\times$3$\to$a4 and falls back to the cheap arm when the next arm's expected wait exceeds a tuned threshold.
The LinUCB control drops both the failure counts and the attempt index.
OctoT2I keeps the routing prompt of its released code, with our arms as its tools and DeepSeek-V4-Flash~\cite{deepseekv4} at temperature 0 as its LLM, and builds its knowledge of each arm from the first-try pass rates of the training folds instead of self-evolution.
It stops at the first pass, after $R$ rounds or once the most expensive arm has failed twice, and \sys is compared with it at a budget equal to its mean cost.

\section{Implementation and Testbed}
\label{app:testbed}
\label{sec:impl}
\sys runs as a router in front of one resident worker per GPU, each holding one editing model or the judge in memory.
The router tracks queue lengths and service-time estimates, passes them to the planner as extra cost, and dispatches the next attempt of a request as soon as the previous one is judged.
Load replay feeds the logged edits, with their verdicts and measured service times, through a discrete-event model of the GPU queues and runs the same planner and policy code as the router.
The judge worker runs EditScore at $\tau=7.5$.
The 160 requests arrive as steady Poisson streams of 0.045 and 0.055\,req/s and as a bursty on--off stream averaging 0.05\,req/s.
The bursty stream switches between 1.6 and 0.4 times its mean rate every 600\,s.
The queue threshold is re-tuned per load in load replay, RouteLLM-MF and HADIS run as per-item sequences trained offline, and load-replay sweeps use 1{,}500 requests and 3 seeds per point.

In four further runs the DP is never behind the best fixed ladder.
On a second, harder request stream at 0.045\,req/s both reach the same goodput, and the DP cuts p95 latency from 520\,s to 320\,s.
It leads by 2.5 points at 0.04\,req/s, significantly by 3.8 points under bursty arrivals on a second machine, and by 1.9 points on a DiffusionDB~\cite{diffusiondb} arrival trace.
At 0.055\,req/s a request waits in queues for 40\,s on average under the DP and for 136\,s under the fixed ladder.
Table~\ref{tab:qth} runs policies tuned at one load at the others in load replay, over 800 requests and 3 seeds per load.

\begin{table}[t]
\caption{Goodput (\%) in load replay when a policy tuned at load L, written @L, runs at other loads.}
\label{tab:qth}
\small
\setlength{\tabcolsep}{3.5pt}
\begin{tabular*}{\linewidth}{@{\extracolsep{\fill}}lccccc}
\toprule
Policy (load in req/s) & 0.04 & 0.05 & 0.06 & burst 0.055 & worst \\
\midrule
Queue thr.\ @0.045 & 80.50 & 77.60 & 49.30 & 61.70 & 49.30 \\
Queue thr.\ @0.055 & 80.50 & 80.10 & 78.80 & 76.30 & 76.30 \\
Queue thr.\ @burst 0.05 & 79.70 & 79.70 & 79.00 & 78.40 & 78.40 \\
Ladder @0.045 & 80.70 & 79.60 & 74.80 & 73.50 & 73.50 \\
Ladder @0.055 & 80.40 & 80.00 & 78.60 & 76.60 & 76.60 \\
Ladder @burst 0.05 & 79.40 & 78.50 & 76.80 & 77.20 & 76.80 \\
\midrule
\textbf{Ours}, one $\lambda$ & \textbf{82.40} & \textbf{81.60} & \textbf{80.00} & \textbf{79.20} & \textbf{79.20} \\
\bottomrule
\end{tabular*}
\end{table}

Over the 30 testbed runs of Figure~\ref{fig:calib}, success differs by at most 2.5 points, the median error of mean latency is $-$1.3\%, and 26 of 30 runs are within 11\% either way.
In four runs, all three RouteLLM-MF runs and the HADIS run at 0.055\,req/s, load replay underestimates mean latency by 12--22\% and p95 by 13--34\%.

A second testbed serves pool~2 with b1, b2 and the judge on one RTX 4090D GPU each and the serve half of GEdit-CN.
Its 160 requests arrive as one Poisson stream at the steady load of Table~\ref{tab:serve} scaled by the ratio of the two pools' expensive arm times (0.0203\,req/s) under a service-level objective of 814\,s scaled the same way, and every knob is picked in load replay as for Table~\ref{tab:serve}.
\sys reaches a goodput of 81.2\%, against 67.5\% for same-model retry and at most 51.2\% for RouteLLM-MF, a two-level cascade and a router in the style of HADIS, with every request that hits the router's 3600\,s timeout counted as failed.

\section{Further Results}
\label{app:more}

\input{tables/tab_human}
\input{tables/tab_octo}
\input{tables/tab_loopjudge}

In the quality gate of \S\ref{sec:judge}, an attempt on pool~2 passes only if the ensemble judge accepts it and its VIEScore~\cite{viescore} perceptual-quality (PQ) score, the smaller of the naturalness and artifact ratings given by Qwen3.8-Flash-Next~\cite{qwen38next} with the original prompt, reaches the bar.
The middle bar is the lower median of LongCat's first-try PQ on the training folds, 6 on GEdit-CN and 7 on GEdit-EN and ImgEdit, and the other two are one point lower and higher.
Methods that read the judge's score see the smaller of the ensemble score and the PQ score shifted so that the two bars coincide.
Every method is refit on the gated labels with the same folds, grid and budget rule, and without the gate the replay reproduces Table~\ref{tab:main} exactly.
Under the gate the strongest prior baseline is the best of the nine that do not rewrite the instruction.
Pool~1 uses the same gate, PQ score and refitting, with a middle bar of 6 on all five sets and the other two one point lower and higher, and its strongest prior baseline is the best of all eleven.
Blind rewrite-retry is regenerated for the requests that the gate newly fails, while critique rewrite-retry is not and enters with a pessimistic and an optimistic bound on its pass rate, under both of which \sys ties the strongest prior baseline at the middle bar.

With a blind rewrite of the cheap arm as one more action (\S\ref{sec:offline}), \sys is significantly above blind rewrite-retry in all nine gated settings at the main budget, including the strictest bar on ImgEdit, where it passes 53.33\% of requests against 48.89\%.

We also charge each judge call 5\,s instead of 10\,s and re-tune every method under the new charge.
At the main budget \sys leads the best of the nine prior baselines that do not rewrite significantly in all eight groups.

The third pool runs FLUX.2 [klein] 9B with eight seeds and Qwen-Image-2.1 and GLM-Image with two seeds each on the 438 GEdit-EN requests, timed on A800 GPUs. GLM-Image conditions on and outputs images whose sides are multiples of 32, with the short side at least 512 and the long side at most 2048 pixels.
Its main budget is the cost of one GLM-Image attempt (37.5\,s), where \sys is 5.9 points above the strongest single arm.

For text-to-image generation we take the 553 GenEval prompts and the five tools of OctoT2I as arms, namely SD-Turbo and SDXL-Turbo~\cite{add}, SANA-Sprint~\cite{sanasprint}, SANA~1.5~\cite{sana15} and Flow-GRPO~\cite{flowgrpo} on SD3.5-Medium~\cite{sd3}, with its NVILA-Lite-2B verifier~\cite{nvila,sana15} as the in-loop judge.
GenEval's six subtasks serve as categories, every method is refit with the folds, grid and budget rule of \S\ref{sec:eval}, and the main budget is the cost of one Flow-GRPO attempt, 6.98\,s.
At the main budget \sys is 2.5 points above two Flow-GRPO attempts at 63\% of their cost.
OctoT2I runs its released loop with the same tools, its own knowledge files and DeepSeek-V4-Flash, and at its mean cost under round limits of two, three and four \sys is significantly more successful, while with a single round OctoT2I costs less than the cheapest setting of \sys.

Table~\ref{tab:human} lists all four judges against the annotators of \S\ref{sec:judge}.
The agreement of the in-loop ensemble is on par with that of the external VIEScore judge, and the ensemble makes the fewest false accepts of the four.
Of 200 sampled attempts, 186 keep a majority vote once unsure labels are dropped, and the annotators agree on 170 of the 180 attempts they both judged (Cohen's kappa 0.856).
A stratum combines a group, the two single judges' verdicts and whether the score is near the pass line. Every group has equal weight, and CIs, also of the paired gaps between judges, bootstrap within strata.
FA and FR are the shares of all attempts that a judge wrongly accepts and wrongly rejects, respectively.

\section{Use of Generative Artificial Intelligence}
\label{app:genai}
Generative artificial intelligence assistants helped draft and edit the text and write the experiment and plotting code.
An image-generation model sketched candidate layouts for Figure~\ref{fig:overview}, which we redrew as a vector diagram, and drew some of its icons.
The authors checked every statement, number and figure against the experiment records and take full responsibility for the paper.

\clearpage
\end{document}

%% file: tables/tab_main.tex
\begin{table*}[t]
\caption{Offline success rate (\%) at the main budget of 50\,s for pool~1 and 120\,s for pool~2.}
\label{tab:main}
\small
\begin{tabular*}{\textwidth}{@{\extracolsep{\fill}}lcccccccc}
\toprule
& \multicolumn{5}{c}{Pool 1 (50\,s)} & \multicolumn{3}{c}{Pool 2 (120\,s)} \\
\cmidrule(lr){2-6}\cmidrule(lr){7-9}
Method & GEdit-CN & GEdit-EN & EN 5 arms & GEdit-v2 & ImgEdit & GEdit-CN & GEdit-EN & ImgEdit \\
\midrule
\color{black!55}Oracle, per-item hindsight & \color{black!55}92.466 & \color{black!55}91.553 & \color{black!55}93.151 & \color{black!55}88.261 & \color{black!55}87.778 & \color{black!55}87.671 & \color{black!55}88.584 & \color{black!55}92.778 \\
\midrule
\multicolumn{9}{l}{\emph{Retry or route one model}} \\
\quad Same-model retry & 81.735 & 81.735 & 82.648 & 80.435 & 78.889 & 77.397 & 73.973 & 79.444 \\
\quad Rewrite-retry, blind & 80.822 & 82.420 & 84.703 & 79.565 & 81.667 & 77.397 & 73.973 & 77.778 \\
\quad Rewrite-retry, critique & 78.311 & 81.279 & 83.333 & 77.391 & 80.000 & 77.397 & 73.973 & 79.444 \\
\quad Best-of-N & 74.886 & 76.256 & 79.224 & 74.348 & 74.444 & 77.397 & 73.973 & 79.444 \\
\quad Cost-Aware Routing~\cite{costaware} & 83.790 & 83.333 & 86.530 & 80.435 & 83.333 & 77.397 & 78.082 & 87.222 \\
\quad RouteLLM-SW~\cite{routellm} & 81.735 & 81.735 & 82.648 & 80.435 & 78.889 & 77.397 & 73.973 & 82.222 \\
\quad RouteLLM-MF~\cite{routellm} & 82.420 & 82.420 & 85.388 & 80.435 & \textbf{85.000} & 77.854 & 76.941 & 83.333 \\
\midrule
\multicolumn{9}{l}{\emph{Cascade in a fixed order}} \\
\quad FrugalGPT~\cite{frugalgpt} & 79.452 & 81.507 & 85.616 & 80.435 & 81.667 & 84.475 & 81.735 & 86.667 \\
\quad AutoMix~\cite{automix} & 81.735 & 83.790 & 86.301 & 83.043 & 81.667 & 84.475 & 82.648 & 86.667 \\
\quad DiffServe~\cite{diffserve} & 84.247 & 83.562 & 87.443 & 82.609 & 83.889 & 82.877 & 82.877 & 86.667 \\
\quad HADIS~\cite{hadis} & 84.247 & 83.562 & 87.443 & 81.739 & 83.333 & 82.877 & 82.877 & 86.667 \\
\midrule
\multicolumn{9}{l}{\emph{Controls without failure history}} \\
\quad \color{black!55}DP planner on first-try rates & \color{black!55}84.018 & \color{black!55}85.388 & \color{black!55}88.356 & \color{black!55}83.478 & \color{black!55}79.444 & \color{black!55}85.388 & \color{black!55}86.301 & \color{black!55}92.778 \\
\quad \color{black!55}LinUCB without retry history, ACQB-like & \color{black!55}86.073 & \color{black!55}84.932 & \color{black!55}86.986 & \color{black!55}83.913 & \color{black!55}85.556 & \color{black!55}77.397 & \color{black!55}74.658 & \color{black!55}90.000 \\
\midrule
\multicolumn{9}{l}{\emph{\sys with failure history}} \\
\quad Fixed ladder & \underline{87.900} & \textbf{88.813} & \textbf{90.639} & \textbf{83.913} & \underline{84.444} & \textbf{85.845} & \underline{85.160} & \textbf{92.778} \\
\quad Online, LinUCB & 87.215 & \underline{84.247} & \underline{87.900} & \underline{83.478} & \textbf{85.000} & 81.507 & 78.539 & \underline{91.667} \\
\quad \textbf{\sys}, DP planner & \textbf{88.813} & \textbf{88.813} & \textbf{90.639} & \textbf{83.913} & \textbf{85.000} & \underline{84.932} & \textbf{85.616} & \textbf{92.778} \\
\bottomrule
\end{tabular*}
\end{table*}

%% file: tables/tab_xfer.tex
\begin{table}[t]
\caption{Success (\%) of Ours and the best prior baseline (Prior) in-domain, on unseen categories and across benchmarks.}
\label{tab:xfer}
\small
\setlength{\tabcolsep}{3.2pt}
\begin{tabular*}{\linewidth}{@{\extracolsep{\fill}}clcccccc}
\toprule
& & \multicolumn{2}{c}{In-domain} & \multicolumn{2}{c}{Unseen category} & \multicolumn{2}{c}{Cross-benchmark} \\
\cmidrule(lr){3-4}\cmidrule(lr){5-6}\cmidrule(lr){7-8}
Pool & Test set & Prior & Ours & Prior & Ours & Prior & Ours \\
\midrule
1 & GEdit-CN & 84.25 & \textbf{88.81} & 84.25 & \textbf{87.67} & 85.16 & \textbf{87.56} \\
1 & GEdit-EN & 83.79 & \textbf{88.81} & 83.79 & \textbf{88.81} & 84.47 & \textbf{88.01} \\
1 & EN 5 arms & 87.44 & \textbf{90.64} & 87.44 & \textbf{90.87} & -- & -- \\
1 & GEdit-v2 & 83.04 & \textbf{83.91} & 82.61 & \textbf{84.35} & 83.26 & \textbf{83.70} \\
1 & ImgEdit & 85.00 & \textbf{85.00} & 83.89 & \textbf{85.56} & 83.89 & \textbf{85.28} \\
\midrule
2 & GEdit-CN & 84.47 & \textbf{84.93} & 84.47 & \textbf{86.07} & 84.47 & \textbf{85.16} \\
2 & GEdit-EN & 82.88 & \textbf{85.62} & 82.88 & \textbf{84.93} & 82.88 & \textbf{84.70} \\
2 & ImgEdit & 87.22 & \textbf{92.78} & 86.67 & \textbf{92.78} & 86.67 & \textbf{92.78} \\
\bottomrule
\end{tabular*}
\end{table}

%% file: tables/tab_serve.tex
\begin{table}[t]
\caption{Testbed goodput (\%) and p95 latency (s) at three loads, with p95 in italics when it exceeds the 300\,s SLO.}
\label{tab:serve}
\small
\setlength{\tabcolsep}{3.2pt}
\begin{tabular*}{\linewidth}{@{\extracolsep{\fill}}lcccccc}
\toprule
& \multicolumn{2}{c}{Steady 0.045} & \multicolumn{2}{c}{Steady 0.055} & \multicolumn{2}{c}{Bursty 0.05} \\
\cmidrule(lr){2-3}\cmidrule(lr){4-5}\cmidrule(lr){6-7}
Method & Good. & p95 & Good. & p95 & Good. & p95 \\
\midrule
Queue thr. & 85.00 & 218.1 & 85.00 & 190.6 & \textbf{84.38} & 210.2 \\
Cascade a5$\to$a4 & 81.25 & 167.4 & 80.00 & 248.4 & 78.75 & 277.5 \\
Fixed ladder & 80.62 & \textit{447.3} & 76.88 & \textit{777.4} & 81.25 & \textit{361.7} \\
HADIS & 76.25 & \textit{631.6} & 75.62 & \textit{945.2} & 75.00 & \textit{1064.3} \\
Retry a5$\times$8 & 73.12 & \textit{869.7} & 70.62 & \textit{1479.8} & 58.13 & \textit{1938.6} \\
RouteLLM-MF & 56.88 & \textit{1629.0} & 56.25 & \textit{1792.8} & 57.50 & \textit{2153.2} \\
\textbf{Ours} & \textbf{86.25} & 279.2 & \textbf{86.25} & 296.3 & \textbf{84.38} & 275.1 \\
\bottomrule
\end{tabular*}
\end{table}

%% file: tables/tab_cells.tex
\begin{table}[t]
\caption{First-try and retry pass rates and their ratio in percent, as ranges over the arms of each group.}
\label{tab:cells}
\small
\setlength{\tabcolsep}{3.5pt}
\begin{tabular*}{\linewidth}{@{\extracolsep{\fill}}lrrrrrr}
\toprule
& \multicolumn{3}{c}{EditScore} & \multicolumn{3}{c}{EditReward} \\
\cmidrule(lr){2-4}\cmidrule(lr){5-7}
Group & first & retry & ratio & first & retry & ratio \\
\midrule
P1 GEdit-CN & 65--73 & 16--30 & 24--41 & 66--75 & 15--22 & 22--29 \\
P1 GEdit-EN & 70--75 & 14--29 & 21--39 & 68--76 & 16--22 & 24--30 \\
P1 EN 5 arms & 48--75 & 14--29 & 21--42 & 41--79 & 11--23 & 24--37 \\
P1 GEdit-v2 & 68--73 & 18--30 & 26--43 & 65--70 & 17--22 & 25--32 \\
P1 ImgEdit & 64--72 & 22--37 & 34--52 & 63--74 & 13--24 & 21--32 \\
\midrule
P2 GEdit-CN & 44--73 & 21--25 & 33--53 & 45--72 & 15--26 & 24--36 \\
P2 GEdit-EN & 43--74 & 21--28 & 37--48 & 46--71 & 11--20 & 20--30 \\
P2 ImgEdit & 66--73 & 23--27 & 34--40 & 63--75 & 11--31 & 17--46 \\
\bottomrule
\end{tabular*}
\end{table}

%% file: tables/tab_human.tex
\begin{table}[t]
\caption{Agreement of each judge with the majority vote of two annotators, reweighted to all attempts, with 95\% CIs.}
\label{tab:human}
\small
\setlength{\tabcolsep}{2.4pt}
\begin{tabular*}{\linewidth}{@{\extracolsep{\fill}}lccccc}
\toprule
Judge & Agreement (\%) & $\kappa$ & AUC & FA (\%) & FR (\%) \\
\midrule
In-loop ensemble & 83.70 [79.95, 87.39] & 0.621 & 0.935 & 2.61 & 13.69 \\
EditScore & 75.93 [72.48, 79.16] & 0.450 & 0.853 & 5.74 & 18.33 \\
EditReward & 78.65 [75.14, 82.05] & 0.512 & 0.890 & 4.37 & 16.97 \\
VIEScore & 79.53 [73.65, 85.25] & 0.520 & 0.853 & 5.04 & 15.43 \\
\bottomrule
\end{tabular*}
\end{table}

%% file: tables/tab_octo.tex
\begin{table}[t]
\caption{Lead of \sys in points over OctoT2I at OctoT2I's own cost for each round limit $R$, with the cost of OctoT2I at $R=1$ in GPU seconds per request.}
\label{tab:octo}
\small
\setlength{\tabcolsep}{3.2pt}
\begin{tabular*}{\linewidth}{@{\extracolsep{\fill}}clccccc}
\toprule
& & & \multicolumn{4}{c}{Round limit $R$} \\ \cmidrule(lr){4-7}
Pool & Test set & Cost & 1 & 2 & 3 & 4 \\
\midrule
1 & GEdit-CN & 60.12 & +17.35 & +10.05 & +7.76 & +6.85 \\
1 & GEdit-EN & 65.49 & +15.98 & +9.82 & +6.39 & +5.02 \\
1 & EN 5 arms & 97.44 & +13.47 & +5.48 & +3.42 & +2.97 \\
1 & GEdit-v2 & 47.18 & +14.78 & +11.74 & +7.39 & +5.65 \\
1 & ImgEdit & 24.04 & +5.56 & +3.33 & $-$0.56 & $-$0.56 \\
\midrule
2 & GEdit-CN & 124.2 & +7.08 & +4.11 & +1.83 & +0.68 \\
2 & GEdit-EN & 123.3 & +13.01 & +5.94 & +2.28 & +1.14 \\
2 & ImgEdit & 91.75 & +14.44 & +10.56 & +5.00 & +4.44 \\
\bottomrule
\end{tabular*}
\end{table}

%% file: tables/tab_loopjudge.tex
\begin{table}[t]
\caption{Success (\%) at the main budget of the strongest prior baseline and of \sys when a single judge replaces the ensemble in the loop.}
\label{tab:loopjudge}
\small
\setlength{\tabcolsep}{3.2pt}
\begin{tabular*}{\linewidth}{@{\extracolsep{\fill}}clcccc}
\toprule
& & \multicolumn{2}{c}{EditScore in loop} & \multicolumn{2}{c}{EditReward in loop} \\ \cmidrule(lr){3-4}\cmidrule(lr){5-6}
Pool & Test set & Prior & Ours & Prior & Ours \\
\midrule
1 & GEdit-CN & 83.56 & 84.93 & 82.65 & 84.70 \\
1 & GEdit-EN & 85.62 & 87.44 & 82.65 & 82.88 \\
1 & EN 5 arms & 85.39 & 87.90 & 84.70 & 87.21 \\
1 & GEdit-v2 & 85.65 & 90.43 & 78.70 & 81.74 \\
1 & ImgEdit & 87.78 & 89.44 & 82.78 & 82.22 \\
\midrule
2 & GEdit-CN & 83.33 & 85.39 & 81.28 & 83.79 \\
2 & GEdit-EN & 83.11 & 86.07 & 79.68 & 81.05 \\
2 & ImgEdit & 85.56 & 89.44 & 83.33 & 86.11 \\
\bottomrule
\end{tabular*}
\end{table}

%% file: refs_gen.bib
@article{frugalgpt,
  title = {{FrugalGPT}: How to Use Large Language Models While Reducing Cost and Improving Performance},
  author = {Chen, Lingjiao and Zaharia, Matei and Zou, James},
  journal = {Transactions on Machine Learning Research},
  year = {2024}
}

@inproceedings{automix,
  title = {{AutoMix}: Automatically Mixing Language Models},
  author = {Pranjal Aggarwal and Aman Madaan and Ankit Anand and others},
  booktitle = {Advances in Neural Information Processing Systems (NeurIPS)},
  pages = {131000--131034},
  year = {2024},
  volume = {37}
}

@inproceedings{routellm,
  title = {{RouteLLM}: Learning to Route {LLMs} from Preference Data},
  author = {Ong, Isaac and Almahairi, Amjad and Wu, Vincent and Chiang, Wei-Lin and Wu, Tianhao and Gonzalez, Joseph E. E. and Kadous, M. Waleed and Stoica, Ion},
  booktitle = {International Conference on Learning Representations (ICLR)},
  year = {2025}
}

@article{costaware,
  title = {Cost-Aware Routing for Efficient Text-to-Image Generation},
  author = {Li, Qinchan and Chen, Kenneth and Su, Changyue and Jitkrittum, Wittawat and Sun, Qi and Sangkloy, Patsorn},
  journal = {Transactions on Machine Learning Research},
  year = {2026}
}

@inproceedings{diffserve,
  title = {{DiffServe}: Efficiently Serving Text-to-Image Diffusion Models with Query-Aware Model Scaling},
  author = {Ahmad, Sohaib and Yang, Qizheng and Wang, Haoliang and Sitaraman, Ramesh K. and Guan, Hui},
  booktitle = {Proceedings of Machine Learning and Systems (MLSys)},
  volume = {7},
  year = {2025}
}

@misc{hadis,
  title = {{HADIS}: A Hybrid Architecture for Query-Aware Diffusion Model Serving},
  author = {Qizheng Yang and Tung-I Chen and Siyu Zhao and Ramesh K. Sitaraman and Hui Guan},
  year = {2025},
  eprint = {2509.00642},
  archivePrefix = {arXiv}
}

@misc{genrouter,
  title = {{GenRouter}: Unified Workflow Routing for Agentic Image Generation},
  author = {Harold Haodong Chen and Zhiyu Hou and Wen-Jie Shu and Weilin Ruan and Yingjie Xu and Litao Guo and Ying-Cong Chen},
  year = {2026},
  eprint = {2608.16721},
  archivePrefix = {arXiv}
}

@inproceedings{octot2i,
  title = {{OctoT2I}: A Self-Evolving Agentic Text-to-Image Router},
  author = {Jiang, Xu and Chen, Bin and Li, Gehui and Duan, Yule and Wang, Ronggang and Zhang, Jian},
  booktitle = {Proceedings of the IEEE/CVF Conference on Computer Vision and Pattern Recognition (CVPR)},
  pages = {31628--31638},
  year = {2026}
}

@misc{deepseekv4,
  title = {{DeepSeek-V4}: Towards Highly Efficient Million-Token Context Intelligence},
  author = {{DeepSeek-AI}},
  year = {2026},
  eprint = {2606.19348},
  archivePrefix = {arXiv}
}

@inproceedings{geneval,
  title = {{GenEval}: An Object-Focused Framework for Evaluating Text-to-Image Alignment},
  author = {Ghosh, Dhruba and Hajishirzi, Hannaneh and Schmidt, Ludwig},
  booktitle = {Advances in Neural Information Processing Systems (NeurIPS)},
  year = {2023}
}

@inproceedings{add,
  title = {Adversarial Diffusion Distillation},
  author = {Sauer, Axel and Lorenz, Dominik and Blattmann, Andreas and Rombach, Robin},
  booktitle = {Proceedings of the European Conference on Computer Vision (ECCV)},
  pages = {87--103},
  year = {2024}
}

@inproceedings{sanasprint,
  title = {{SANA-Sprint}: One-Step Diffusion with Continuous-Time Consistency Distillation},
  author = {Chen, Junsong and Xue, Shuchen and Zhao, Yuyang and Yu, Jincheng and Paul, Sayak and Chen, Junyu and Cai, Han and Han, Song and Xie, Enze},
  booktitle = {Proceedings of the IEEE/CVF International Conference on Computer Vision (ICCV)},
  pages = {16185--16195},
  year = {2025}
}

@inproceedings{sana15,
  title = {{SANA} 1.5: Efficient Scaling of Training-Time and Inference-Time Compute in Linear Diffusion Transformer},
  author = {Xie, Enze and Chen, Junsong and Zhao, Yuyang and others},
  booktitle = {International Conference on Machine Learning (ICML)},
  volume = {267},
  pages = {68578--68598},
  year = {2025}
}

@inproceedings{flowgrpo,
  title = {Flow-{GRPO}: Training Flow Matching Models via Online {RL}},
  author = {Liu, Jie and Liu, Gongye and Liang, Jiajun and Li, Yangguang and Liu, Jiaheng and Wang, Xintao and Wan, Pengfei and Zhang, Di and Ouyang, Wanli},
  booktitle = {Advances in Neural Information Processing Systems (NeurIPS)},
  year = {2025}
}

@inproceedings{sd3,
  title = {Scaling Rectified Flow Transformers for High-Resolution Image Synthesis},
  author = {Esser, Patrick and Kulal, Sumith and Blattmann, Andreas and others},
  booktitle = {International Conference on Machine Learning (ICML)},
  volume = {235},
  pages = {12606--12633},
  year = {2024}
}

@inproceedings{nvila,
  title = {{NVILA}: Efficient Frontier Visual Language Models},
  author = {Liu, Zhijian and Zhu, Ligeng and Shi, Baifeng and others},
  booktitle = {Proceedings of the IEEE/CVF Conference on Computer Vision and Pattern Recognition (CVPR)},
  pages = {4122--4134},
  year = {2025}
}

@inproceedings{treacle,
  title = {Efficient Contextual {LLM} Cascades Through Budget-Constrained Policy Learning},
  author = {Zhang, Xuechen and Huang, Zijian and Taga, Ege Onur and Joe-Wong, Carlee and Oymak, Samet and Chen, Jiasi},
  booktitle = {Advances in Neural Information Processing Systems (NeurIPS)},
  pages = {91691--91722},
  year = {2024},
  volume = {37}
}

@misc{acqb,
  title = {Learning to Route and Schedule {LLMs} from User Retrials via Contextual Queueing Bandits},
  author = {Seoungbin Bae and Junyoung Son and Dabeen Lee},
  year = {2026},
  eprint = {2602.02061},
  archivePrefix = {arXiv}
}

@misc{resample,
  title = {Resample or Reroute? Recoverable Stopping Debt Without Identified Action Selection},
  author = {Teng-Ruei Chen},
  year = {2026},
  eprint = {2607.08665},
  archivePrefix = {arXiv}
}

@misc{latencyroute,
  title = {Mind the Delay: Latency-aware Query Routing for Heterogeneous {LLM} pools},
  author = {Shivam Patel and Akaash R. Parthasarathy and Akash Dhasade and Ankur Mallick and Gauri Joshi},
  year = {2026},
  eprint = {2607.18253},
  archivePrefix = {arXiv}
}

@inproceedings{editscore,
  title = {{EditScore}: Unlocking Online {RL} for Image Editing via High-Fidelity Reward Modeling},
  author = {Luo, Xin and Wang, Jiahao and Wu, Chenyuan and Xiao, Shitao and Jiang, Xiyan and Lian, Defu and Zhang, Jiajun and Liu, Dong and Liu, Zheng},
  booktitle = {International Conference on Learning Representations (ICLR)},
  year = {2026}
}

@inproceedings{editreward,
  title = {{EditReward}: A Human-Aligned Reward Model for Instruction-Guided Image Editing},
  author = {Wu, Keming and Jiang, Sicong and Ku, Max and Nie, Ping and Liu, Minghao and Chen, Wenhu},
  booktitle = {International Conference on Learning Representations (ICLR)},
  year = {2026}
}

@inproceedings{imgedit,
  title = {{ImgEdit}: A Unified Image Editing Dataset and Benchmark},
  author = {Yang Ye and Xianyi He and Zongjian Li and Bin Lin and Shenghai Yuan and Zhiyuan Yan and Bohan Hou and Li Yuan},
  booktitle = {Advances in Neural Information Processing Systems (NeurIPS)},
  pages = {146002--146032},
  year = {2025},
  volume = {38}
}

@misc{geditv2,
  title = {{GEditBench} v2: A Human-Aligned Benchmark for General Image Editing},
  author = {Zhangqi Jiang and Zheng Sun and Xianfang Zeng and Yufeng Yang and Xuanyang Zhang and Yongliang Wu and Wei Cheng and Gang Yu and Xu Yang and Bihan Wen},
  year = {2026},
  eprint = {2603.28547},
  archivePrefix = {arXiv}
}

@misc{step1xedit,
  title = {{Step1X-Edit}: A Practical Framework for General Image Editing},
  author = {Shiyu Liu and Yucheng Han and Peng Xing and others},
  year = {2025},
  eprint = {2504.17761},
  archivePrefix = {arXiv}
}

@misc{reasonedit,
  title = {{ReasonEdit}: Towards Reasoning-Enhanced Image Editing Models},
  author = {Fukun Yin and Shiyu Liu and Yucheng Han and others},
  year = {2025},
  eprint = {2511.22625},
  archivePrefix = {arXiv}
}

@misc{qwenimage,
  title = {Qwen-Image Technical Report},
  author = {Chenfei Wu and Jiahao Li and Jingren Zhou and others},
  year = {2025},
  eprint = {2508.02324},
  archivePrefix = {arXiv}
}

@misc{kontext,
  title = {{FLUX.1} Kontext: Flow Matching for In-Context Image Generation and Editing in Latent Space},
  author = {{Black Forest Labs} and others},
  year = {2025},
  eprint = {2506.15742},
  archivePrefix = {arXiv}
}

@misc{vibe,
  title = {{VIBE}: Visual Instruction Based Editor},
  author = {Grigorii Alekseenko and Aleksandr Gordeev and Irina Tolstykh and Bulat Suleimanov and Vladimir Dokholyan and Georgii Fedorov and Sergey Yakubson and Aleksandra Tsybina and Mikhail Chernyshov and Maksim Kuprashevich},
  year = {2026},
  eprint = {2601.02242},
  archivePrefix = {arXiv}
}

@misc{longcat,
  title = {{LongCat-Image} Technical Report},
  author = {{Meituan LongCat Team} and Hanghang Ma and Haoxian Tan and others},
  year = {2025},
  eprint = {2512.07584},
  archivePrefix = {arXiv}
}

@misc{qwen3vl,
  title = {{Qwen3-VL} Technical Report},
  author = {Shuai Bai and Yuxuan Cai and Ruizhe Chen and others},
  year = {2025},
  eprint = {2511.21631},
  archivePrefix = {arXiv}
}

@misc{qwen25vl,
  title = {{Qwen2.5-VL} Technical Report},
  author = {Shuai Bai and Keqin Chen and Xuejing Liu and others},
  year = {2025},
  eprint = {2502.13923},
  archivePrefix = {arXiv}
}

@misc{vllmsr,
  title = {{vLLM} Semantic Router: Signal Driven Decision Routing for Mixture-of-Modality Models},
  author = {Xunzhuo Liu and Huamin Chen and Samzong Lu and others},
  year = {2026},
  eprint = {2603.04444},
  archivePrefix = {arXiv}
}

@inproceedings{diffusiondb,
  title = {{DiffusionDB}: A Large-scale Prompt Gallery Dataset for Text-to-Image Generative Models},
  author = {Wang, Zijie J. and Montoya, Evan and Munechika, David and Yang, Haoyang and Hoover, Benjamin and Chau, Duen Horng},
  booktitle = {Proceedings of the 61st Annual Meeting of the Association for Computational Linguistics (ACL)},
  pages = {893--911},
  year = {2023}
}

@inproceedings{selfrefine,
  title = {Self-Refine: Iterative Refinement with Self-Feedback},
  author = {Madaan, Aman and Tandon, Niket and Gupta, Prakhar and others},
  booktitle = {Advances in Neural Information Processing Systems (NeurIPS)},
  pages = {46534--46594},
  year = {2023},
  volume = {36}
}

@inproceedings{reflexion,
  title = {Reflexion: Language Agents with Verbal Reinforcement Learning},
  author = {Shinn, Noah and Cassano, Federico and Gopinath, Ashwin and Narasimhan, Karthik and Yao, Shunyu},
  booktitle = {Advances in Neural Information Processing Systems (NeurIPS)},
  pages = {8634--8652},
  year = {2023},
  volume = {36}
}

@inproceedings{genartist,
  title = {{GenArtist}: Multimodal {LLM} as an Agent for Unified Image Generation and Editing},
  author = {Wang, Zhenyu and Li, Aoxue and Li, Zhenguo and Liu, Xihui},
  booktitle = {Advances in Neural Information Processing Systems (NeurIPS)},
  pages = {128374--128395},
  year = {2024},
  volume = {37}
}

@inproceedings{idea2img,
  title = {{Idea2Img}: Iterative Self-refinement with {GPT-4V} for Automatic Image Design and Generation},
  author = {Yang, Zhengyuan and Wang, Jianfeng and Li, Linjie and Lin, Kevin and Lin, Chung-Ching and Liu, Zicheng and Wang, Lijuan},
  booktitle = {Proceedings of the European Conference on Computer Vision (ECCV)},
  pages = {167--184},
  year = {2024}
}

@inproceedings{linucb,
  title = {A Contextual-Bandit Approach to Personalized News Article Recommendation},
  author = {Li, Lihong and Chu, Wei and Langford, John and Schapire, Robert E.},
  booktitle = {Proceedings of the 19th International Conference on World Wide Web (WWW)},
  pages = {661--670},
  year = {2010}
}

@article{simonkadane,
  title = {Optimal Problem-Solving Search: All-or-None Solutions},
  author = {Simon, Herbert A. and Kadane, Joseph B.},
  journal = {Artificial Intelligence},
  volume = {6},
  number = {3},
  pages = {235--247},
  year = {1975}
}

@article{bellman,
  title = {On the Theory of Dynamic Programming},
  author = {Bellman, Richard},
  journal = {Proceedings of the National Academy of Sciences},
  volume = {38},
  number = {8},
  pages = {716--719},
  year = {1952}
}

@article{gittins,
  title = {Bandit Processes and Dynamic Allocation Indices},
  author = {Gittins, John C.},
  journal = {Journal of the Royal Statistical Society Series B: Statistical Methodology},
  volume = {41},
  number = {2},
  pages = {148--164},
  year = {1979}
}

@article{weitzman,
  title = {Optimal Search for the Best Alternative},
  author = {Weitzman, Martin L.},
  journal = {Econometrica},
  volume = {47},
  number = {3},
  pages = {641--654},
  year = {1979}
}

@article{efron79,
  title = {Bootstrap Methods: Another Look at the Jackknife},
  author = {Efron, Bradley},
  journal = {The Annals of Statistics},
  volume = {7},
  number = {1},
  year = {1979},
  pages = {1--26}
}

@article{efronmorris,
  title = {Data Analysis Using Stein's Estimator and Its Generalizations},
  author = {Efron, Bradley and Morris, Carl},
  journal = {{Journal of the American Statistical Association}},
  volume = {70},
  number = {350},
  pages = {311--319},
  year = {1975}
}

@article{little,
  title = {A Proof for the Queuing Formula: $L = \lambda W$},
  author = {Little, John D. C.},
  journal = {Operations Research},
  volume = {9},
  number = {3},
  pages = {383--387},
  year = {1961}
}

@incollection{rice,
  title = {The Algorithm Selection Problem},
  author = {Rice, John R.},
  booktitle = {Advances in Computers},
  pages = {65--118},
  year = {1976},
  volume = {15}
}

@inproceedings{violajones,
  title = {Rapid Object Detection Using a Boosted Cascade of Simple Features},
  author = {Viola, Paul and Jones, Michael},
  booktitle = {Proceedings of the IEEE Computer Society Conference on Computer Vision and Pattern Recognition (CVPR)},
  volume = {1},
  pages = {511--518},
  year = {2001}
}

@article{auer,
  title = {Finite-time Analysis of the Multiarmed Bandit Problem},
  author = {Auer, Peter and Cesa-Bianchi, Nicolò and Fischer, Paul},
  journal = {Machine Learning},
  volume = {47},
  number = {2--3},
  pages = {235--256},
  year = {2002}
}

@inproceedings{distserve,
  title = {{DistServe}: Disaggregating Prefill and Decoding for Goodput-optimized Large Language Model Serving},
  author = {Yinmin Zhong and Shengyu Liu and Junda Chen and Jianbo Hu and Yibo Zhu and Xuanzhe Liu and Xin Jin and Hao Zhang},
  booktitle = {18th USENIX Symposium on Operating Systems Design and Implementation (OSDI)},
  year = {2024},
  pages = {193--210}
}

@inproceedings{ldm,
  title = {High-Resolution Image Synthesis with Latent Diffusion Models},
  author = {Rombach, Robin and Blattmann, Andreas and Lorenz, Dominik and Esser, Patrick and Ommer, Björn},
  booktitle = {Proceedings of the IEEE/CVF Conference on Computer Vision and Pattern Recognition (CVPR)},
  pages = {10674--10685},
  year = {2022}
}

@inproceedings{progdistill,
  title = {Progressive Distillation for Fast Sampling of Diffusion Models},
  author = {Salimans, Tim and Ho, Jonathan},
  booktitle = {International Conference on Learning Representations (ICLR)},
  year = {2022}
}

@inproceedings{instructpix2pix,
  title = {{InstructPix2Pix}: Learning to Follow Image Editing Instructions},
  author = {Brooks, Tim and Holynski, Aleksander and Efros, Alexei A.},
  booktitle = {Proceedings of the IEEE/CVF Conference on Computer Vision and Pattern Recognition (CVPR)},
  pages = {18392--18402},
  year = {2023}
}

@inproceedings{viescore,
  title = {{VIEScore}: Towards Explainable Metrics for Conditional Image Synthesis Evaluation},
  author = {Ku, Max and Jiang, Dongfu and Wei, Cong and Yue, Xiang and Chen, Wenhu},
  booktitle = {Proceedings of the 62nd Annual Meeting of the Association for Computational Linguistics (Volume 1: Long Papers)},
  pages = {12268--12290},
  year = {2024}
}

@inproceedings{llmjudge,
  title = {Judging {LLM-as-a-Judge} with {MT-Bench} and Chatbot Arena},
  author = {Zheng, Lianmin and Chiang, Wei-Lin and Sheng, Ying and others},
  booktitle = {Advances in Neural Information Processing Systems (NeurIPS)},
  pages = {46595--46623},
  year = {2023},
  volume = {36}
}

@inproceedings{react,
  title = {{ReAct}: Synergizing Reasoning and Acting in Language Models},
  author = {Yao, Shunyu and Zhao, Jeffrey and Yu, Dian and Du, Nan and Shafran, Izhak and Narasimhan, Karthik R. and Cao, Yuan},
  booktitle = {International Conference on Learning Representations (ICLR)},
  year = {2023}
}

@inproceedings{edival,
  title = {{EdiVal-Agent}: An Object-Centric Framework for Automated, Fine-Grained Evaluation of Multi-Turn Editing},
  author = {Chen, Tianyu and Zhang, Yasi and Zhang, Zhi and others},
  booktitle = {International Conference on Learning Representations (ICLR)},
  year = {2026}
}

@inproceedings{risebench,
  title = {Envisioning Beyond the Pixels: Benchmarking Reasoning-Informed Visual Editing},
  author = {Zhao, Xiangyu and Zhang, Peiyuan and Tang, Kexian and others},
  booktitle = {Advances in Neural Information Processing Systems (NeurIPS)},
  pages = {192865--192904},
  year = {2025},
  volume = {38}
}

@inproceedings{jarvisevo,
  title = {{JarvisEvo}: Towards a Self-Evolving Photo Editing Agent with Synergistic Editor-Evaluator Optimization},
  author = {Lin, Yunlong and Wang, Linqing and Lin, Kunjie and others},
  booktitle = {Proceedings of the IEEE/CVF Conference on Computer Vision and Pattern Recognition (CVPR)},
  pages = {27291--27302},
  year = {2026}
}


%% file: refs_manual.bib
@misc{flux2klein,
  title = {{FLUX.2 [klein] 4B}},
  author = {{Black Forest Labs}},
  year = {2026},
  note = {Hugging Face model card},
  howpublished = {\url{https://huggingface.co/black-forest-labs/FLUX.2-klein-4B}}
}

@misc{flux2klein9b,
  title = {{FLUX.2 [klein] 9B}},
  author = {{Black Forest Labs}},
  year = {2026},
  note = {Hugging Face model card},
  howpublished = {\url{https://huggingface.co/black-forest-labs/FLUX.2-klein-9B}}
}

@misc{glmimage,
  title = {{GLM-Image}},
  author = {{Z.ai}},
  year = {2026},
  note = {Hugging Face model card},
  howpublished = {\url{https://huggingface.co/zai-org/GLM-Image}}
}

@misc{turboalpha,
  title = {{FLUX.1-Turbo-Alpha}},
  author = {{Alimama Creative}},
  year = {2024},
  note = {Hugging Face model card},
  howpublished = {\url{https://huggingface.co/alimama-creative/FLUX.1-Turbo-Alpha}}
}

@misc{lightning,
  title = {{Qwen-Image-Edit-2511-Lightning}},
  author = {{LightX2V}},
  year = {2025},
  note = {Hugging Face model card},
  howpublished = {\url{https://huggingface.co/lightx2v/Qwen-Image-Edit-2511-Lightning}}
}

@misc{qwen38next,
  title = {{Qwen3.8-Flash-Next}: A New Architecture, Towards Ultimate Cost-Efficiency},
  author = {{Qwen Team}},
  year = {2026},
  howpublished = {\url{https://qwen.ai/blog?id=qwen3.8-flash-next}}
}

@misc{qwenimage21,
  title = {{Qwen-Image-2.1}},
  author = {{Qwen Team}},
  year = {2026},
  note = {Hugging Face model card},
  howpublished = {\url{https://huggingface.co/Qwen/Qwen-Image-2.1}}
}
